\documentclass[a4paper,fleqn]{cas-dc}

\usepackage[numbers,sort&compress]{natbib}
\usepackage[normalem]{ulem}
\usepackage{hyperref}

\usepackage{flushend}

\hypersetup{
    colorlinks=true,
    linkcolor=blue,
    citecolor=blue
}

\usepackage{caption}
\usepackage{cuted}
\usepackage{algorithm}
\usepackage{float}
\usepackage{algpseudocode}

\def\tsc#1{\csdef{#1}{\textsc{\lowercase{#1}}\xspace}}
\tsc{WGM}
\tsc{QE}
\tsc{EP}
\tsc{PMS}
\tsc{BEC}
\tsc{DE}
\AtBeginDocument{
    \setlength{\abovedisplayskip}{8pt}
    \setlength{\belowdisplayskip}{8pt}
    \setlength{\abovedisplayshortskip}{8pt}
    \setlength{\belowdisplayshortskip}{8pt}
    \setlength{\mathindent}{0pt}
}

\begin{document}
\let\WriteBookmarks\relax
\def\floatpagepagefraction{1}
\def\textpagefraction{.001}
\shorttitle{Multi-agent Driven Formal Instruction Generation Framework}
\shortauthors{Shixing Zhao et~al.}

\title [mode = title]{MAFIG: Multi-agent Driven Formal Instruction Generation Framework}

\author[1]{Shixing Zhao}[orcid=0009-0001-5894-5133]
\ead{shixing@gs.zzu.edu.cn}

\author[1]{Zheng Si}
% [orcid=0009-0008-2382-7891]
\author[1]{Pengpeng Ouyang}
% [orcid=0009-0003-3104-6247]
\author[1]{Zhengqing Hu}
% [orcid=0009-0008-7293-2735]
\author[1]{Wanqi Zhu}
% [orcid=0009-0004-4964-2335]
\author[1,2,3,4]{Dong Chen}[orcid=0000-0002-4859-1757]
\ead{chendongai@zzu.edu.cn}
\cormark[1]
\author[1,2,3,4]{Yibo Guo}[orcid=0000-0002-7336-6252]
\ead{ieybguo@zzu.edu.cn}
\cormark[1]
\author[1,2,3,4]{Mingliang Xu}
\ead{iexumingliang@zzu.edu.cn}
\cormark[1]

\affiliation[1]{organization={School of Computer and Artificial Intelligence, Zhengzhou University},
                city={Zhengzhou},
                postcode={450001},
                state={Henan},
                country={China}}

\affiliation[2]{organization={Engineering Research Center of Intelligent Swarm Systems, Ministry of Education},
                city={Zhengzhou},
                postcode={450001},
                state={Henan},
                country={China}}

\affiliation[3]{organization={National Supercomputing Center in Zhengzhou},
                city={Zhengzhou},
                postcode={450001},
                state={Henan},
                country={China}}

\affiliation[4]{organization={Henan Research Center for Large Model Technology and New Quality Software Engineering},
                city={Zhengzhou},
                postcode={450001},
                state={Henan},
                country={China}}                     
\cortext[1]{Corresponding author}

\begin{abstract}
Emergency situations in scheduling systems often trigger local functional failures that undermine system stability and even cause system collapse. Existing methods primarily rely on robust scheduling or reactive scheduling, handling emergencies through predefined rules or rescheduling strategies. However, the diversity and unpredictability of real-world emergencies make them difficult to anticipate, which limits the adaptability of these methods in complex scenarios. Recent studies have shown that Large Language Models (LLMs) possess strong potential for complex scheduling tasks because of their extensive prior knowledge and strong reasoning capabilities. Nevertheless, the high inference latency of LLMs and the lengthy contextual information of scheduling systems significantly hinder their application for emergency handling. To mitigate these issues, we propose the Multi-agent Driven Formal Instruction Generation Framework (MAFIG). The framework constrains the decision scope to local functional modules affected by emergency situations and repairs scheduling logic rapidly by generating formal instructions. MAFIG contains a Perception Agent and an Emergency Decision Agent, which mitigates the adverse impact of lengthy system contexts on emergency decision-making. We further introduce span-focused loss-driven local distillation mechanism (SFL) to transfer the decision-making capability of powerful Cloud Large Language Models (C-LLMs) to lightweight local models, reducing inference latency while preserving decision-making effectiveness. Experiments in the Port, Warehousing, and Deck scheduling datasets show success rates of 98.49\%, 94.97\%, and 97.50\%, with average processing times of 0.33 s, 0.23 s, and 0.19 s. These results demonstrate that MAFIG effectively mitigates the impact of emergencies and improves the robustness and adaptability of scheduling systems.
\end{abstract}

\maketitle

\section{Introduction}
Modern scheduling systems are essential for ensuring the orderly and efficient execution of tasks in industrial scenarios \cite{chen2023task,agnetis2025fifty}, such as ports and warehousing. Most existing scheduling systems are built on tightly coupled architectures, which achieve satisfactory performance in static and mildly dynamic environments  \cite{ouelhadj2009survey}. However, when environmental states, system resources, or scheduling constraints fluctuate, local emergency situations can easily trigger cascading failures across the system  \cite{lu2025hybrid}, ultimately undermining the continuous and stable operation. Consequently, improving the robustness and sustained operability of scheduling systems requires decoupling complex architectures into functionally distinct and relatively independent submodules  \cite{liu2024reconfigurable}, enabling localized correction under emergency situations.

\begin{figure}
    \centering
    \includegraphics[width=\columnwidth]{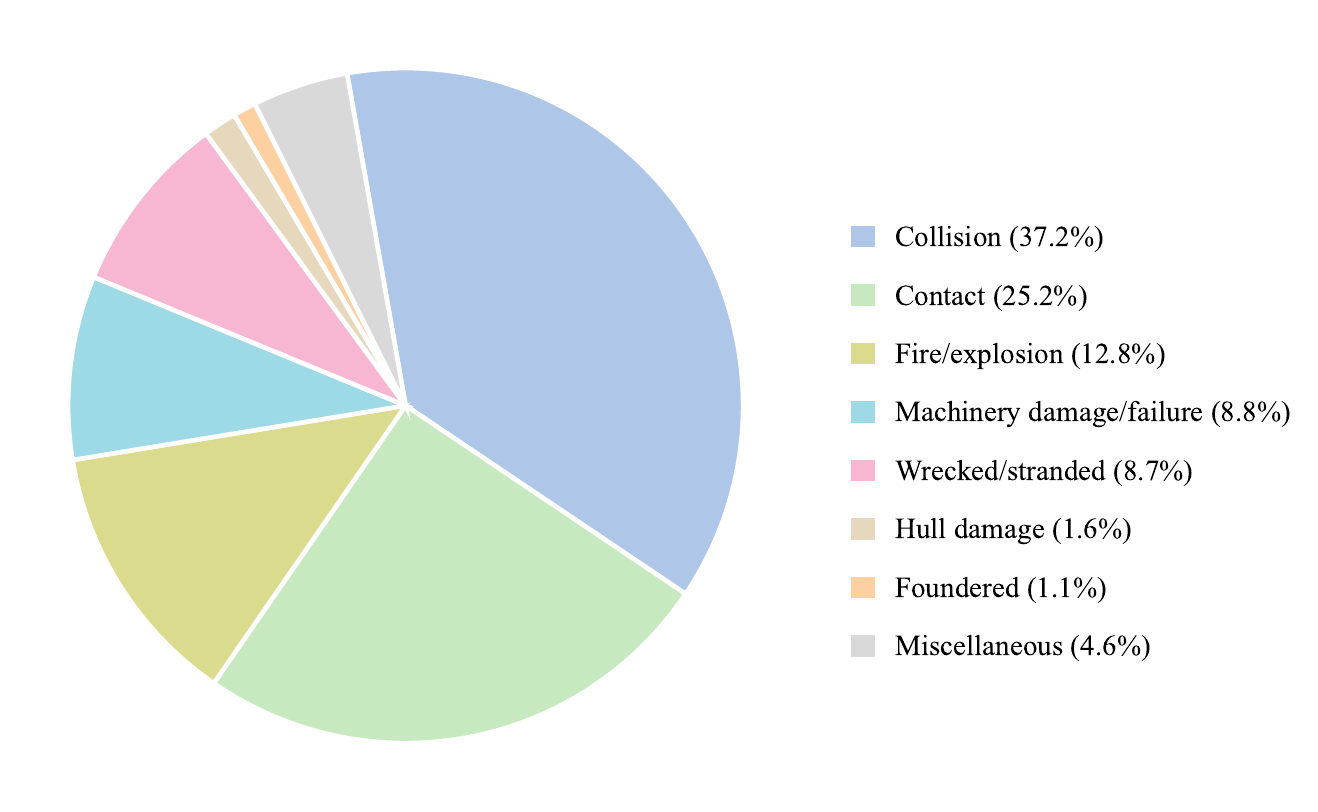}
    \caption{Accident types in Mediterranean port areas.}
    \label{FIG:1}
\end{figure}

By leveraging task decomposition and allocation mechanisms, existing decoupled scheduling systems decompose complex scheduling problems into relatively independent subtasks and employ efficient evaluation frameworks \cite{framinan2010architecture,sadeh1998blackboard,smith1994reactive}, thereby improving planning effectiveness and operational responsiveness. However, in practical operations, although emergency situations occur with relatively low frequency, they often significantly disrupt the overall scheduling plan \cite{ouelhadj2009survey}. As illustrated in \hyperref[FIG:1]{Fig.~\ref*{FIG:1}}, empirical statistics on Mediterranean port accidents \cite{marino2023analysis} indicate that emergency situations are characterized by a broad variety and severe impacts, which make them inherently difficult to anticipate, limiting the ability of scheduling systems to handle them through predefined rules. Meanwhile, the severe consequences of emergencies require timely and appropriate response measures to mitigate their impact on system operations. For instance, vessel arrival delays may be triggered by heterogeneous factors such as adverse weather or channel control. These delays do not merely postpone a single operational node. Instead, they propagate across multiple functional modules, including berth allocation, quay crane scheduling, and yard planning, and leading to cascading scheduling conflicts and systemic resource underutilization. Nevertheless, most existing decoupled scheduling systems are still built on static or semi-static modeling assumptions, which limit their resilience to emergency situations \cite{ghaleb2020real}. When emergency situations cause functional modules to fail, these systems often lack efficient mechanisms for revising their core logical structure \cite{herroelen2005project}. Although recent studies have introduced adaptive scheduling and optimization modules \cite{vieira2003adaptive}, such adjustments are typically limited to parameter or policy levels and cannot directly modify the core logical structure of the scheduling system.

In recent years, the rapid advancement of Large Language Models (LLMs) in natural language understanding, logical reasoning, and code generation has drawn increasing attention to their potential in complex planning and scheduling problems \cite{abgaryan2024llms,tang2025kct}. First, owing to their strong language understanding capabilities \cite{li2025retrieval,chen2025logic,brahmachary2025leo}, LLMs can perform high-level semantic parsing of input instructions and convert unstructured or incomplete information into actionable inputs for decision-making. Second, when confronted with emergency situations, LLMs do not need to rely on manually specified rules. Instead, by parsing the instructions and integrating with their internal extensive prior knowledge and strong reasoning capabilities \cite{brown2020gpt3,wei2022cot,kojima2022zeroshot}, they can generate feasible formal instructions for handling emergency situations. Finally, with their code generation and modification capabilities, LLMs demonstrate the potential to directly revise the core logical structure of scheduling systems, rather than remaining limited to parameter or policy adjustments. Despite these advantages, the direct integration of LLMs into scheduling systems for emergency handling still faces two major challenges: (1) \textbf{Latency-Quality Tradeoff}: model size is generally positively correlated with decision-making capability, but increasing the number of parameters inevitably leads to a substantial rise in inference latency \cite{chen2025improving}. In scheduling systems driven by large models, delayed responses to emergencies may further amplify their operational consequences. (2) \textbf{Execution Misalignment}: existing decoupled scheduling systems often exhibit complex architectures and lengthy module dependency chains. When emergencies occur, it is difficult to rapidly identify the affected functional modules, which increases the risk that the decisions generated by LLMs may fail to align with the core logic of the scheduling system.

Building on the above analysis, we propose the Multi-agent Driven Formal Instruction Generation Framework (MAFIG). Under emergency situations, MAFIG no longer performs reasoning and global modification over the entire scheduling system. Instead, it constrains its scope to the affected local functional modules, transfers the corresponding atomic functions into Formal Instructions, and achieves rapid repair through targeted modification of relevant atomic functions or the generation of new atomic functions. Specifically, MAFIG consists of a Perception Agent and an Emergency Decision Agent that operate collaboratively. To address the Latency-Quality Tradeoff, MAFIG introduces the span-focused loss-driven local distillation mechanism (SFL) to transfer the decision-making capability of powerful Cloud Large Language Models (C-LLMs) to lightweight local models. As a result, MAFIG substantially reduces inference latency while preserving decision-making effectiveness, satisfying the real-time requirements of highly dynamic and time-sensitive scheduling scenarios. To mitigate Execution Misalignment, MAFIG explicitly formalizes the processes of emergency perception, impact analysis, and function localization. The Perception Agent is responsible for the semantic parsing of emergency situations and the analysis of their impact. If emergency situations have a substantive impact on the current scheduling plan, the Perception Agent rapidly identifies the affected atomic functions and transfers them, along with the relevant emergency information, to the Emergency Decision Agent. Based on this information, the Emergency Decision Agent revises existing atomic functions or generates new ones, and then writes them into the atomic function library. Through this mechanism, MAFIG maintains continuous system operation and progressively improves its efficiency and adaptability in responding to similar emergency situations.

The main contributions of this paper are summarized as follows: \begin{enumerate}[(1)]
\item We analyze the limitations of existing decoupled scheduling systems in handling emergency situations, with particular emphasis on their limited ability to revise the core logical structure when functional modules fail. Furthermore, we identify two key challenges in the direct integration of Large Language Models into scheduling systems: Latency-Quality Tradeoff and Execution Misalignment.
\item We propose the Multi-agent Driven Formal Instruction Generation Framework (MAFIG) for handling emergencies in scheduling systems. By combining collaborative multi-agent decision-making with the span-focused loss-driven local distillation mechanism (SFL), MAFIG enables rapid local repair of scheduling logic under emergency situations while reducing inference latency and preserving decision-making effectiveness.
\item We conduct systematic simulation experiments in three representative scenarios: Port, Warehouse, and Deck scheduling. The experimental results demonstrate that MAFIG significantly enhances the stability and adaptability of scheduling systems under diverse emergency situations.
\end{enumerate}

\section{Related work}

\subsection{Emergency Scheduling Systems}
To mitigate functional failures in scheduling systems caused by emergency situations, existing studies can generally be classified into two categories: robust scheduling and reactive scheduling. Robust scheduling mitigates the impact of potential emergency situations by estimating the probability and impact of possible disturbances, incorporating time and resource buffers into the baseline schedule \cite{demeulemeester2009robust}. However, it is often criticized for being overly conservative, since protecting against events with a low probability usually comes at the expense of the initial objective value, such as a significant increase in the overall project makespan, which may be economically unjustifiable \cite{portoleau2020robust}. In contrast, reactive scheduling repairs the baseline schedule after emergency situations occur. It adapts the current plan to new conditions through local modification or global rescheduling. Its main advantage lies in strong task specificity, since it does not require sacrificing system efficiency in advance. Nevertheless, delayed responses may allow their effects to spread rapidly across the scheduling network. Frequent rescheduling may also impose a considerable computational burden on the system \cite{herroelen2004robust}.
Despite these efforts, existing approaches still suffer from a fundamental limitation. Emergencies in real-world scenarios are highly diverse, which makes it difficult for the original scheduling systems to fully account for potential emergency situations at the design stage \cite{chai2018traffic,jedrzejowicz2014rl}. Consequently, relying on static buffering mechanisms or conventional rescheduling strategies is often insufficient to achieve efficient and stable system recovery in complex emergency environments. These limitations further motivate the study of intelligent emergency decision-making for complex scheduling systems, especially approaches that combine the powerful capability of LLMs with multi-agent collaboration.

\subsection{Application of LLMs in scheduling systems}
With the rapid development of artificial intelligence, particularly in natural language processing (NLP) \cite{wen2025acugpt}, Large Language Models (LLMs) have demonstrated strong reasoning and decision-making capabilities in code generation, code completion, and code repair \cite{chen2024data,chen2025kka,huang2026emmamba}. Code-oriented LLMs, exemplified by Claude Code, Codex, and Code Llama, can already generate executable programs from natural language descriptions. They have also achieved strong performance across multiple code benchmarks \cite{chen2021codex,roziere2023codellama}. These advances open a new technical pathway for handling emergency situations in complex scheduling systems. LLMs can parse emergency descriptions and participate in the modification of scheduling logic and code-level reconfiguration. Recent studies in scheduling further indicate that, through task-specific data construction followed by fine-tuning, LLMs can be applied to combinatorial optimization problems such as job shop scheduling, where they show promising reasoning capability in dynamic scheduling scenarios \cite{abgaryan2025starjob,an2026jshm}.

The application of LLMs in decoupled scheduling systems still faces critical challenges. Complex scheduling systems usually involve highly coupled functions, lengthy system contexts, and strict resource constraints. If code-oriented LLMs are directly used for code modification without task-specific adaptation or fine-tuning, they are easily distracted by long contexts, leading to insufficient use of critical information, biased understanding of local logic and myopic decision-making \cite{cao2025reflecsched}. Recent studies further show that code-oriented LLMs may produce code hallucinations, namely code that appears plausible but is actually incorrect, including syntactic errors, logical errors, security vulnerabilities, and even erroneous modifications to existing programs \cite{agarwal2024codemirage}. Consequently, in strongly constrained and highly coupled scheduling systems, modification with code-oriented LLMs is inherently risky. Without explicit constraints on domain rules, function structure, and edit boundaries, models may modify irrelevant code and undermine overall system consistency and executability. To mitigate these challenges, MAFIG precisely localizes the functional modules affected by emergency situations, which reduces interference from irrelevant context during model reasoning. It further introduces explicit domain rules and edit boundary constraints to ensure that the generated repair code remains strictly aligned with the core logic of the scheduling system.

\begin{figure*}
	\centering
	\includegraphics[width=\textwidth]{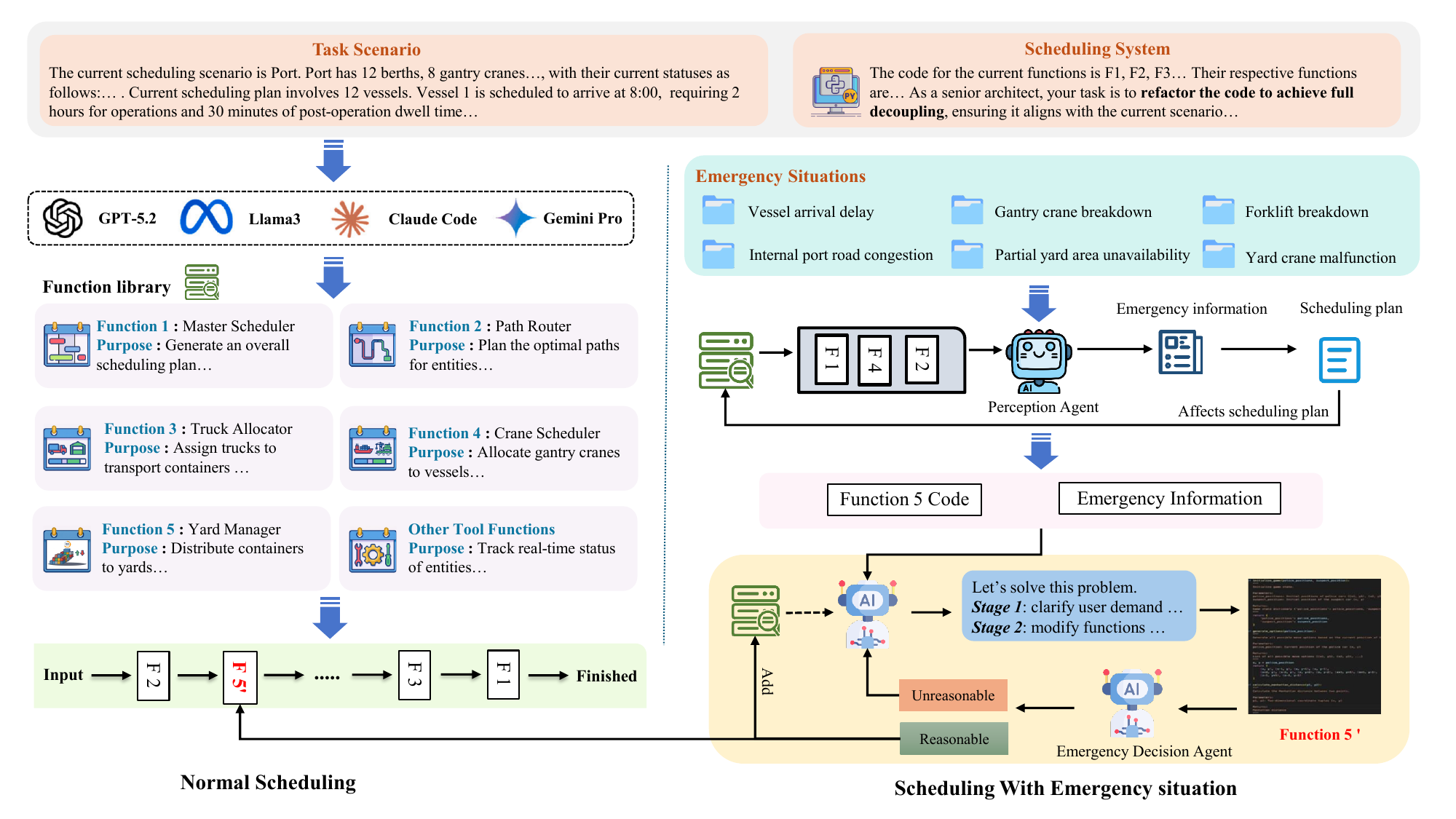}
	\caption{Architecture of MAFIG for emergency decision in scheduling systems. The framework consists of the Perception Agent, the Emergency Decision Agent and the Atomic Function Library. It supports semantic parsing, impact analysis, affected function localization and atomic function revision, which enables rapid recovery of scheduling logic under emergency situations.}
	\label{FIG:2}
\end{figure*}

\section{Methodology}
\subsection{Framework Overview}
As illustrated in \hyperref[FIG:2]{Fig.~\ref*{FIG:2}}, MAFIG primarily consists of three components: the Perception Agent, the Emergency Decision Agent, and the atomic function library. To clearly describe how the framework handles emergency situations in complex and highly dynamic scheduling scenarios, the core variables involved are uniformly defined in \hyperref[tbl1]{Table~\ref*{tbl1}}.

Constrained by the stringent response speed requirements imposed by emergency situations $e_t$, MAFIG employs the span-focused loss-driven local distillation mechanism (SFL) to train efficient lightweight local models that instantiate the Perception Agent $A_P$ and the Emergency Decision Agent $A_E$. When emergency situations $e_t$ occur, $A_P$ acquires the current system state $s_t$, the emergency situations $e_t$, and the function specifications $f_s$. It performs semantic parsing and impact analysis of $e_t$ to localize the affected functional modules $\mathcal{F}_{aff}$ within the atomic function library. Subsequently, $A_P$ forwards $\mathcal{F}_{aff}$, along with the contextual data of $e_t$, to the Emergency Decision Agent $A_E$. Based on this information, $A_E$ revises the relevant atomic functions in $\mathcal{F}_{aff}$ or generates new atomic functions when necessary. Through this process, MAFIG enables localized functional updates in the decoupled scheduling system under emergency situations, which supports rapid repair of scheduling logic and sustained stable system operation.

\begin{table}[width=\linewidth,pos=t] % 改为 pos=t，让它浮动到单栏的顶部
\caption{Description of variables and their corresponding meanings.}\label{tbl1}
\renewcommand{\arraystretch}{1.2} 
% l 代表左对齐，p{0.7\linewidth} 限制右侧描述列的宽度，适应单栏
\begin{tabular*}{\tblwidth}{@{} l p{0.8\linewidth} @{}}
\toprule
Variable & Description \\
\midrule
$t$                 & Current time step \\
$s_t$               & Scheduling environment state at time $t$ \\
$e_t$               & Emergency situations at time $t$ \\
$x$                 & Aggregated system context (system rules, module specifications, code context) \\
$f_s$               & Function specifications \\
$\mathcal{D}_A$     & Atomic function library \\
$\mathcal{F}_{aff}$ & Atomic functions affected by $e_t$ \\
$A_P$               & Perception Agent \\
$A_E$               & Emergency Decision Agent \\
\bottomrule
\end{tabular*}
\end{table}

\subsection{Atomic Function Library}
To address the high coupling in traditional scheduling systems, we employ C-LLMs to perform modular decoupling of complex scheduling logic. Given the aggregated system context $x$, the C-LLMs autoregressively generate the atomic functions of each functional module and the corresponding function specifications $f_s$. The generation process can be formally expressed as follows:
\begin{equation}
    p_{\theta_C}(f, f_s | x) = \prod_{i} p_{\theta_C}(y_i | x, y_{1:i-1}) ,
\end{equation}
where $\theta_C$ denotes the parameters of the C-LLMs, and C-LLMs predict the current token $y_i$ on the basis of the input context and the preceding ${i-1}$ tokens. The generated token sequence $\{y_i\}$ ultimately constitutes an independent atomic function $f$ and the corresponding function specifications $f_s$. The resulting atomic function library is defined as:
\begin{equation}
    \mathcal{D}_A = \{f_m\}_{m=1}^M + f_s,
\end{equation}
where $M$ denotes the number of atomic functions. The atomic function library constructed through this process decomposes the complex scheduling system into a set of functionally explicit and relatively independent submodules. This decomposition provides a standardized interface that enables the Perception Agent $A_P$ to precisely localize $\mathcal{F}_{aff}$ and allows the Emergency Decision Agent $A_E$ to execute targeted modifications.
 
\subsection{Perception Agent}
Although existing Large Language Models exhibit strong decision-making capability in processing complex scheduling logic, their application for emergency handling still faces significant limitations. First, the inference latency of powerful C-LLMs is too high to satisfy the real-time requirements of emergency response in highly dynamic scenarios, which leads to the Latency-Quality Tradeoff. Second, scheduling systems involve highly sensitive core data, so frequent interaction with cloud interfaces introduces severe privacy and security risks. As a consequence, we construct the Perception Agent $A_P$ with a lightweight local model, and then employ distillation to transfer the emergency understanding and reasoning capabilities of the C-LLMs to $A_P$. When emergency situations $e_t$ occur, the Perception Agent $A_P$ performs semantic parsing and impact analysis of $e_t$ to rapidly localize the affected functional modules. This process improves the alignment between subsequent LLM decisions and the core logic of the scheduling system, thereby reducing the risk of Execution Misalignment. In this process, the input to the Perception Agent is defined as: 
\begin{equation}
    z_t = g(e_t, s_t, f_s),
\end{equation}
where $g(\cdot)$ denotes the feature aggregation function, which integrates the emergency situations $e_t$, the current environmental state $s_t$, and the function specifications $f_s$ into a unified input representation.

Conditioned on the input $z_t$, the Perception Agent $A_P$ computes the probability distribution over the atomic functions in $\mathcal{D}_A$ that are affected by the emergency situations. This process establishes a mapping from the semantic space of the emergency situations to the corresponding code modules: 
\begin{equation}
    \mathcal{F}_{aff} = \{ f \in \mathcal{D}_A \mid p_{\theta_P}(f \mid z_t) > \tau \},
\end{equation}
where $\theta_P$ denotes the parameters of the Perception Agent $A_P$, $p_{\theta_P}(f \mid z_t)$ denotes the predicted probability that the atomic function $f$ is affected under the input $z_t$, and $\tau$ represents the decision threshold.

To overcome the limited localization capability caused by the small parameter scale of lightweight local models, we construct a high-quality emergency situations function localization dataset for distillation training. The Perception Agent $A_P$ is optimized with the standard supervised cross-entropy loss:
\begin{equation}
    \mathcal{L}_{AP}(\theta_P) = - \sum_{f \in \mathcal{D}_A} y_P(f) \log p_{\theta_P}(f \mid z_t),
\end{equation}
where $y_P(f)$ denotes the supervision label provided by the C-LLMs to indicate whether the atomic function $f$ is affected. $p_{\theta_P}(f \mid z_t)$ denotes the probability predicted by the Perception Agent that the atomic function $f$ is affected under the input $z_t$.

By minimizing this cross-entropy loss, the Perception Agent $A_P$ inherits the precise localization capability of the C-LLMs, while retaining the efficiency and privacy of local deployment. The resulting set of affected atomic functions $\mathcal{F}_{aff}$ and the relevant emergency information are transferred to the Emergency Decision Agent $A_E$, which provides the basis for subsequent code repair.

\subsection{Emergency Decision Agent}

Precise localization of the affected atomic functions $\mathcal{F}_{aff}$ by the Perception Agent $A_P$ provides the foundation for repairing the affected core logic of the scheduling system. Compared with the semantic parsing and impact analysis performed by $A_P$, the Emergency Decision Agent $A_E$ is responsible for code generation and logic reconstruction. Its training and inference are therefore substantially more demanding.

Constrained by the limited computational resources available in local deployment, the Emergency Decision Agent $A_E$ generally has a relatively small parameter scale. If the distillation stage requires $A_E$ to learn global reconstruction over lengthy functional module code, it is likely to waste its limited fitting capacity on unchanged background code, which leads to long-context forgetting and logical hallucinations. To mitigate this issue, we propose the span-focused loss-driven local distillation mechanism (SFL). While preserving the coherence of the complete code context, SFL assigns differentiated loss weights across code spans, which prevents the model from overfitting irrelevant background information, focusing the limited parameter capacity of $A_E$ and generation capability on the targeted modification of critical code.

Specifically, to train an efficient Emergency Decision Agent $A_E$, we construct the distillation dataset with a diff-based strategy. First, we use the C-LLMs with strong reasoning capability as the teacher model. Conditioned on the integrated context $z_t$ from the previous stage and the atomic function $f \in \mathcal{F}_{aff}$ to be revised, the teacher model generates the target function $f^*$ for handling the emergency situations:
\begin{equation}
    f^* = \text{C-LLMs}({z}_t, f),
\end{equation}
Subsequently, we apply a diff-based strategy to compare the original function $f$ with the target function $f^*$, extract the modified code span between them. Let $[k, k+m]$ denote the index span of the modified segment in $f^*$, which means that the subsequence satisfies $f^*_{k:k+m} \notin f$. To construct supervised training data with explicit edit boundaries, we insert the special tokens \verb|<<EDIT_START>>| and \verb|<<EDIT_END>>| around the modified segment, constructing the final supervisory target sequence $y$:
\begin{equation}
\begin{aligned}
y ={}& f^*_{1:k-1}
\oplus \texttt{<<EDIT\_START>>}
\oplus f^*_{k:k+m} \\
& \oplus \texttt{<<EDIT\_END>>}
\oplus f^*_{k+m+1:|f^*|},
\end{aligned}
\end{equation}
where $\oplus$ denotes the sequence concatenation operation and $|f^*|$ denotes the total length of the target function. This formulation preserves the necessary code context while imposing explicit semantic focus on the modified region $f^*_{k:k+m}$. The resulting sequence $y$ is used as the supervision label for distillation training of the Emergency Decision Agent $A_E$. 

To further alleviate attention dispersion caused by the limited parameter scale of lightweight local models, we design a span-weighted cross-entropy loss for the code editing task. We define the region enclosed by \texttt{<<EDIT\_START>>} and \texttt{<<EDIT\_END>>} in each training sample as the edit span. For an input sequence $x = \{x_1, \dots, x_T\}$, we define a weight vector $\omega \in \mathbb{R}^T$ as follows:
\begin{equation}
    w_i = \begin{cases} 
\lambda_{\text{edit}} & \text{if } x_i \in \text{Edit Span} \\ 
1 & \text{if } x_i \in \text{Normal Context} \\ 
0 & \text{if } x_i \in \text{Padding} 
\end{cases},
\end{equation}
where $\lambda_{\text{edit}}$ is a hyperparameter, usually greater than $1$, that reinforces learning of the edit region. The Emergency Decision Agent $A_E$ is optimized by minimizing the weighted negative log likelihood loss:
\begin{equation}
    \mathcal{L}_{MAFIG}(\theta_E) = - \frac{1}{\sum_{i=1}^{T} w_i} \left( \sum_{i=1}^{T} w_i \log P(x_i \mid x_{<i}; \theta_E) \right)
\end{equation}
A larger $\lambda_{\text{edit}}$ drives gradient updates toward modifications of the core logic of the atomic function, rather than routine syntactic patterns in the surrounding code. This design allows the Emergency Decision Agent $A_E$ to preserve high-quality code generation capability for handling complex emergency situations under a limited parameter scale.

When the revised atomic functions or newly generated atomic functions pass system trial execution, they are added to the atomic function library:
\begin{equation}
\mathcal{D}_A = \mathcal{D}_A \cup f_{\text{new}}.
\end{equation}

Continuous updating of the atomic function library not only supports rapid repair of scheduling logic under emergency situations, but also endows the scheduling system with self-evolution capability. As a result, the system can exhibit stronger robustness and adaptability when similar emergency situations arise in the future.

Furthermore, to enable the model to accurately recognize the introduced special tokens, we extend the tokenizer by adding \verb|<<EDIT_START>>| and \verb|<<EDIT_END>>| to the vocabulary and expand the embedding matrix accordingly. However, direct random initialization of the embeddings for newly introduced tokens may disturb the original semantic space of the pretrained model and destabilize gradients during the early stage of training. To avoid this issue, we adopt a statistics-based initialization strategy. Let $W_{emb} \in \mathbb{R}^{V \times d}$ denote the original embedding matrix. We compute the mean $\mu_{emb}$ and variance $\sigma_{emb}^2$ of the existing token embeddings, then initialize the embedding vector $v_{new}$ of the newly introduced token $t_{new}$ as:
\begin{equation}
    v_{new} = \mu_{emb} + \epsilon, \quad \epsilon \sim \mathcal{N}(0, \gamma \cdot \sigma_{emb}^2),
\end{equation}
where $\gamma$ denotes a scaling factor and $\epsilon$ denotes a minor random perturbation. This initialization makes the newly introduced token embeddings statistically aligned with the existing embedding distribution, which reduces disturbance to the pretrained semantic space and improves optimization stability during the early stage of training.

In summary, we introduces the atomic function library, the Perception Agent, and the Emergency Decision Agent. To provide a clear algorithmic description of MAFIG, we summarize its overall procedure in \hyperref[alg:mafig]{Algorithm~\ref*{alg:mafig}}. 

\begin{algorithm}
\caption{Procedure of MAFIG}
\label{alg:mafig}
\begin{algorithmic}[1]
\Require Aggregated system context $x$, scheduling environment state $s_t$, emergency situations $e_t$, Cloud Large Language Models (C-LLMs), and decision threshold $\tau$
\Ensure Updated atomic function library $\mathcal{D}_A$ for handling emergency situations

\State \textbf{Atomic Function Library Construction:}
\State Use C-LLMs to decouple the scheduling system into atomic functions.
\State Construct the atomic function library $\mathcal{D}_A$ and the function specifications $f_s$.

\State \textbf{Perception Agent Training:}
\State Construct the emergency situation localization dataset.
\State Train the Perception Agent $A_P$.

\State \textbf{Emergency Decision Agent Training:}
\State Construct the distillation dataset.
\State Extend the tokenizer with \texttt{<<EDIT\_START>>} and \texttt{<<EDIT\_END>>}.
\State Train the Emergency Decision Agent $A_E$.

\State \textbf{Perception Agent Inference:}
\State $z_t \gets g(e_t, s_t, f_s)$.
\State Compute $p_{\theta_P}(f \mid z_t)$ for each $f \in \mathcal{D}_A$.
\State $\mathcal{F}_{aff} \gets \{f \in \mathcal{D}_A \mid p_{\theta_P}(f \mid z_t) > \tau\}$.
\State Transfer $\mathcal{F}_{aff}$ and the relevant emergency information to $A_E$.

\State \textbf{Emergency Decision Agent Inference:}
\ForAll{$f \in \mathcal{F}_{aff}$}
\State Generate a revised function or new atomic function $f_{\text{new}}$ with $A_E$.
\State Validate $f_{\text{new}}$ through system trial execution.
\If{$f_{\text{new}}$ passes validation}
\State $\mathcal{D}_A \gets \mathcal{D}_A \cup \{f_{\text{new}}\}$.
\EndIf
\EndFor

\State \Return $\mathcal{D}_A$
\end{algorithmic}
\end{algorithm}

\section{Experiments}
To systematically evaluate the proposed method in complex scheduling scenarios, we construct three simulation datasets by combining the operating mechanisms of real scheduling systems with representative patterns of emergency situations, namely port scheduling dataset EvalPort, warehouse scheduling dataset EvalWare and deck scheduling dataset EvalDeck. These datasets present progressively increasing levels of scenario dynamics, interaction complexity, and constraint stringency. The evaluation is conducted from five perspectives. (1) Overall emergency decision performance of MAFIG across scenarios of increasing complexity. (2) Capacity of MAFIG to jointly mitigate the Latency-Quality Tradeoff and Execution Misalignment. (3) Performance of the MAFIG architecture across different lightweight local models. (4) Effectiveness of SFL. (5) Advantages of SFL over conventional distillation methods in emergency decision tasks.

\subsection{Datasets}
\textbf{EvalPort}. Port terminals play a critical role in world trade as primary nodes connecting sea and land transportation. Berth allocation is a core operation in terminal operations and directly affects subsequent resource allocation, including yard management, equipment deployment, and workforce scheduling. For example, emergency situations can continuously disrupt berth allocation and quay crane scheduling, which increases the demand for dynamic rescheduling in port operations \cite{rodrigues2022berth}. On this basis, we construct EvalPort, which targets core tasks in port scheduling such as berth allocation and quay crane operations. The dataset contains a scheduling function library with eight core atomic functions, 30 localization samples for the Perception Agent and 80 distillation samples for the Emergency Decision Agent. As shown in \hyperref[FIG:3]{Fig.~\ref*{FIG:3}}, the test set covers five categories of emergency situations and contains 199 test instances.

\begin{center}
	\includegraphics[scale=0.4]{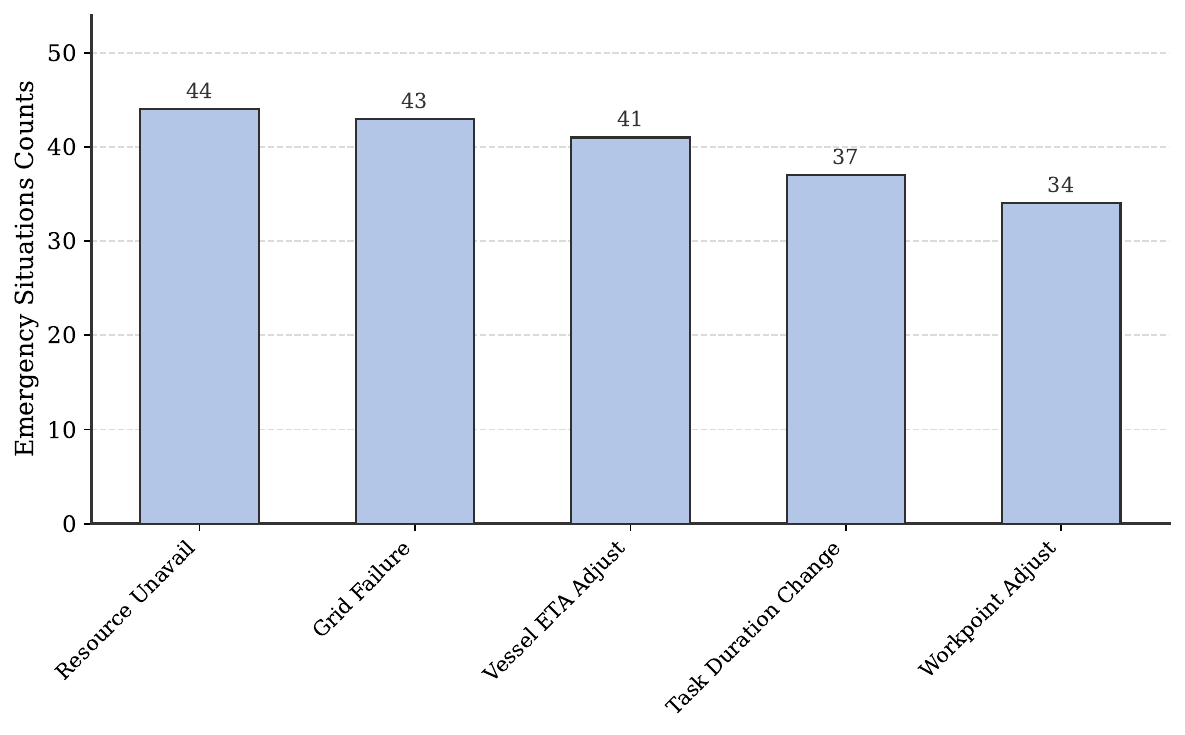}
	\captionof{figure}{Overview of the EvalPort evaluation dataset.}
	\label{FIG:3}
\end{center}

\textbf{EvalWare}. Compared with port scheduling scenarios, warehouse scheduling involves more frequent interactions among operational entities, denser task dependencies, and more intricate propagation paths of local disturbances. These characteristics impose higher requirements on the local replanning capability of the model. Accordingly, EvalWare is constructed for warehouse scheduling scenarios, which covers path planning, dynamic storage location allocation, and coordinated execution of multiple tasks. The dataset contains a scheduling function library composed of 15 core atomic functions, 50 localization samples for the Perception Agent and 170 distillation samples for the Emergency Decision Agent. As shown in \hyperref[FIG:4]{Fig.~\ref*{FIG:4}}, the test set contains 398 emergency instances drawn from eight categories of emergency situations.

\begin{center}
	\includegraphics[width=.5\textwidth]{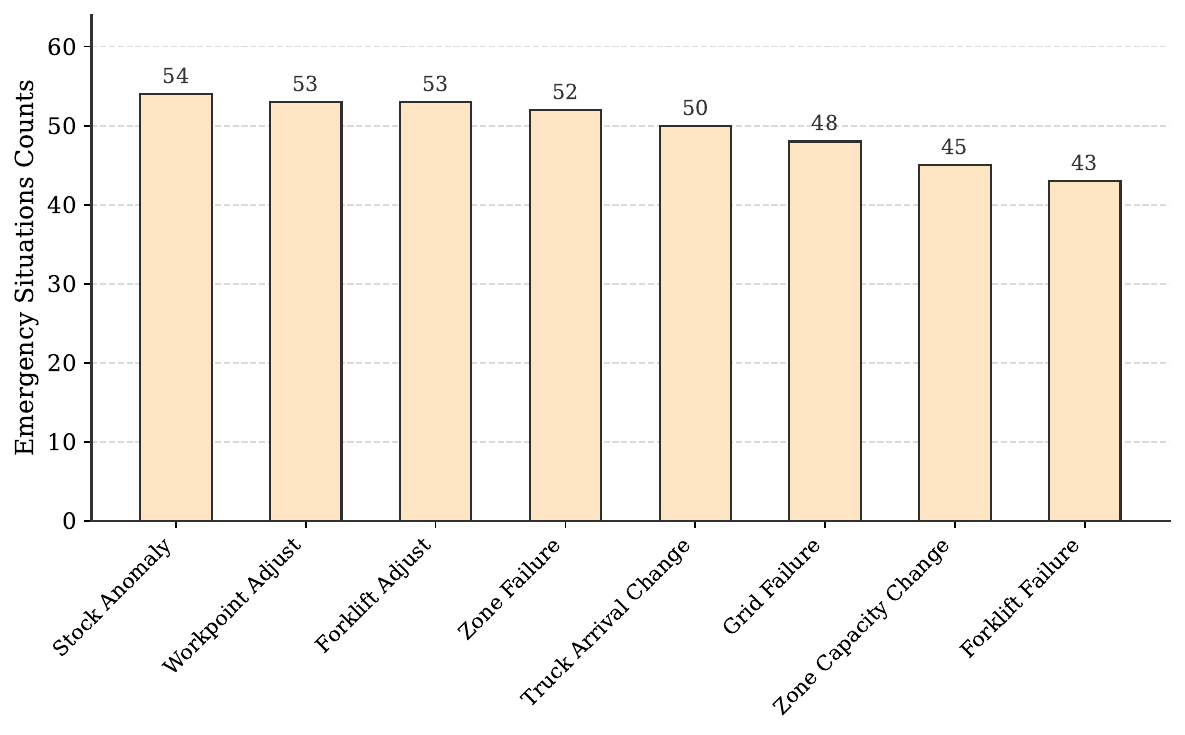}
	\captionof{figure}{Overview of the EvalWare evaluation dataset.}
	\label{FIG:4}
\end{center}

\begin{figure*}
	\centering
	\includegraphics[width=.9\textwidth]{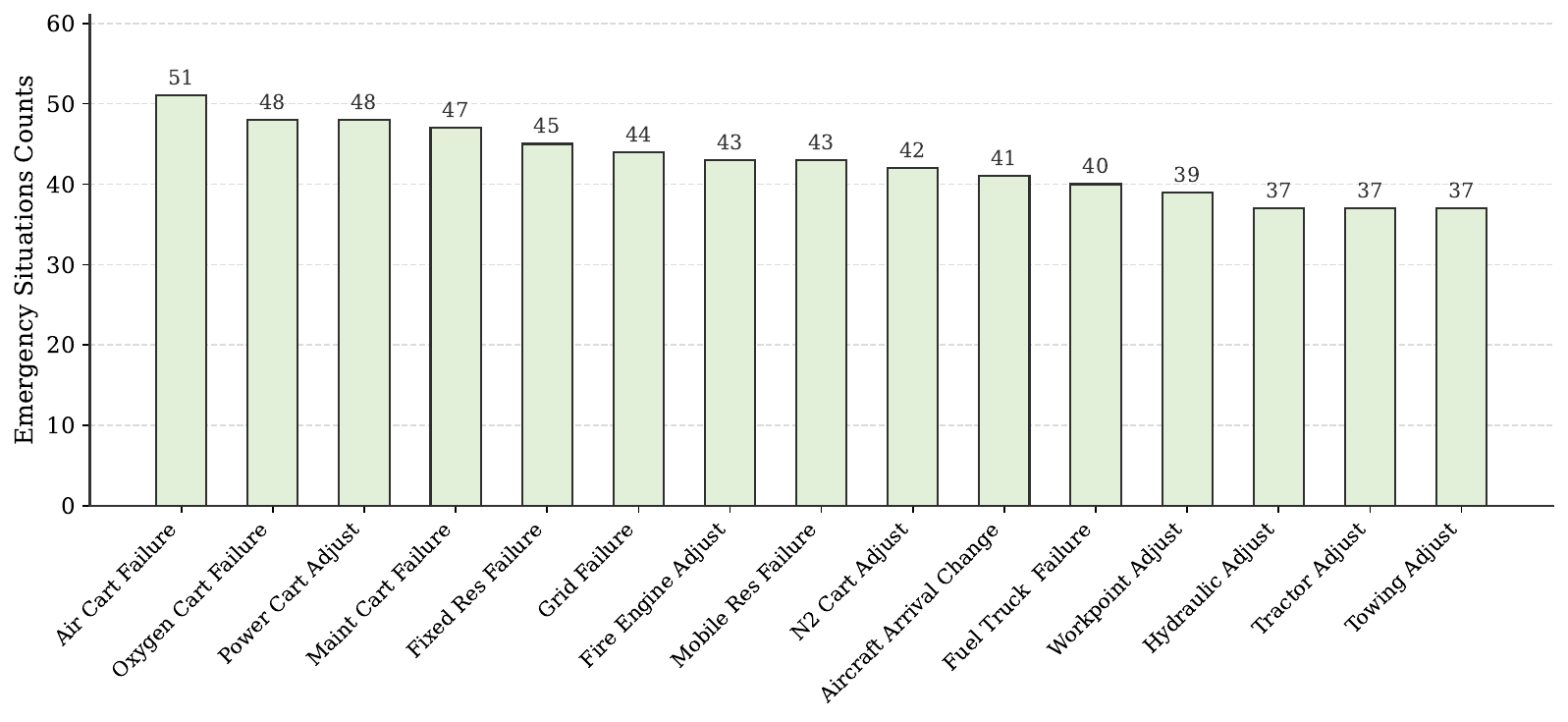}
	\caption{Overview of the EvalDeck evaluation datasets.}
	\label{FIG:5}
\end{figure*}

\textbf{EvalDeck}. Carrier-based aircraft deck scheduling and support operations involve complex path planning, resource allocation, and temporal coordination. This scenario is characterized by severe space limitations, tightly coupled constraints and dynamic disturbances \cite{wang2020review}, which make it the most structurally complex of the three scheduling scenarios. Accordingly, EvalDeck is constructed with a scheduling function library consisting of 25 core atomic functions, 100 localization samples for the Perception Agent, and 120 distillation samples for the Emergency Decision Agent. As illustrated in \hyperref[FIG:5]{Fig.~\ref*{FIG:5}}, the test set contains 642 emergency instances spanning 15 categories of emergency situations, which provides broader coverage of highly complex reconfiguration scenarios.

In summary, EvalPort, EvalWare, and EvalDeck constitute a hierarchical evaluation benchmark for emergency handling in complex scheduling systems, with scenario dynamics, interaction complexity, and constraint intensity increasing progressively across the three datasets. This benchmark enables systematic evaluation of MAFIG in terms of function localization, code repair, and scheduling reconstruction under different levels of scenario complexity.

\subsection{Experiment Setup}
During the distillation phase, SFL is used to enable efficient model distillation. Training is conducted in FP16 precision and the learning rate is initialized to $5 \times 10^{-5}$. The model is trained for 3 epochs with a batch size of 4 and a LoRA rank of 8. AdamW is adopted as the optimizer, and the learning rate is controlled by a cosine scheduler.

During the inference phase, Cloud Large Language Models (C-LLMs) are accessed through cloud APIs, with the generation parameters set to temperature = 0.9 and top-p = 0.95. Lightweight local models are deployed in a local GPU environment using vLLM. All experiments are conducted on a single NVIDIA RTX A6000 GPU. For local inference, the maximum generation length is set to 2560 tokens, and GPU memory utilization is set to 0.9.

\subsection{Experimental Results}
To systematically validate the effectiveness of MAFIG in practical emergency scheduling scenarios, we compare MAFIG, using Qwen2.5-Coder-7B as the backbone model, against four large language models, namely Qwen3-32B \cite{qwen2025qwen3}, Qwen3-Coder-480B-A35B-Instruct, DeepSeek-V3.2 \cite{deepseek2025v32}, and GLM-4.7 \cite{zeng2024chatglm}. The experiments are conducted in port, warehouse, and deck scheduling scenarios, using total task completion time, average processing time, and task success rate as the evaluation metrics. The corresponding results are summarized in \hyperref[tab:performance]{Table~\ref*{tab:performance}}.

Overall, as scenario complexity and constraint intensity increase, most large models exhibit a marked decline in task success rate while still incurring substantial inference latency, which limits their suitability for emergency scheduling scenarios with stringent real-time response requirements. Specifically, GLM-4.7 achieves the best performance among all baselines, with success rates of 89.95\% in the port scenario and 86.18\% in the warehouse scenario. This result reflects the advantage of extensive prior knowledge and strong reasoning capability in complex scheduling tasks. However, its average processing times still reach 27.44 s and 40.75 s. These results indicate that Large Language Models still suffer from the pronounced Latency-Quality Tradeoff in scheduling tasks. Moreover, large-scale models are generally difficult to deploy directly in local scheduling systems and must therefore be accessed through cloud APIs in practical applications. The communication overhead introduced by remote invocation further increases overall response latency. In scheduling scenarios, when emergency situations are not handled promptly, local disturbances tend to propagate and amplify, which makes stable system operation difficult to sustain. Therefore, relying solely on C-LLMs cannot simultaneously satisfy the dual requirements of high-quality decision-making and rapid response in complex scheduling systems. In contrast, MAFIG requires only 0.33 s and 0.23 s on average in the port and warehouse scenarios, while achieving success rates of 98.49\% and 94.97\%, respectively. It thus outperforms C-LLMs in both decision quality and response efficiency. These results show that MAFIG can substantially reduce inference latency while maintaining a higher task success rate, thereby effectively alleviating the Latency-Quality Tradeoff in emergency scheduling tasks. This advantage arises from the proposed SFL, which enables the lightweight local model to inherit effective knowledge and decision patterns from C-LLMs for emergency handling. As a result, the lightweight local model preserves high inference efficiency while retaining the decision capability required for emergency handling.
\begin{table*}[htbp]
    \centering
    \caption{Performance comparison of different models across tasks.}
    \label{tab:performance}
    
    \begin{tabular*}{\textwidth}{@{\extracolsep{\fill}} c l c c c @{}}
        \toprule
        \textbf{Task} & \textbf{Models} & \textbf{Total Time (s)} & \textbf{Avg Time (s)} & \textbf{Success Rate} \\
        \midrule
        
        \multirow{6}{*}{Port} 
        & Qwen3-32B & 1763.73 & 8.86 & 62.31\% \\
        & Qwen3-Coder-480B-A35B-Instruct & 896.53 & 4.51 & 85.93\% \\
        & DeepSeek-V3.2 & 1827.28 & 9.18 & 86.43\% \\
        & GLM-4.7 & 5460.58 & 27.44 & 89.95\% \\
        & Qwen2.5-Coder-7B & 69.20 & 0.35 & 25.62\% \\
        & \textbf{MAFIG(Qwen2.5-Coder-7B)} & \textbf{64.98} & \textbf{0.33} & \textbf{98.49\%} \\
        \cmidrule{2-5} % 横线只从第2列画到第5列，宽度自动适配列范围
        
        % Warehousing 部分
        \multirow{6}{*}{Warehousing} 
        & Qwen3-32B & 6716.86 & 16.88 & 75.38\% \\
        & Qwen3-Coder-480B-A35B-Instruct & 1657.91 & 4.17 & 83.17\% \\
        & DeepSeek-V3.2 & 5134.66 & 12.90 & 83.67\% \\
        & GLM-4.7 & 16217.18 & 40.75 & 86.18\% \\
        & Qwen2.5-Coder-7B & 151.18 & 0.38 & 25.87\% \\
        & \textbf{MAFIG(Qwen2.5-Coder-7B)} & \textbf{90.27} & \textbf{0.23} & \textbf{94.97\%} \\
        \cmidrule{2-5}
        
        % Deck 部分
        \multirow{6}{*}{Deck} 
        & Qwen3-32B & 8300.60 & 12.93 & 50.62\% \\
        & Qwen3-Coder-480B-A35B-Instruct & 2646.40 & 4.12 & 51.40\% \\
        & DeepSeek-V3.2 & 8753.61 & 13.63 & 50.93\% \\
        & GLM-4.7 & 20582.86 & 32.06 & 83.49\% \\
        & Qwen2.5-Coder-7B & 250.13 & 0.39 & 17.45\% \\
        & \textbf{MAFIG(Qwen2.5-Coder-7B)} & \textbf{119.87} & \textbf{0.19} & \textbf{97.50\%} \\
        
        \bottomrule
    \end{tabular*}
\end{table*}

Moreover, with the growth of scenario complexity and constraint intensity, Execution Misalignment during C-LLMs inference becomes increasingly evident. This issue is particularly severe in the deck scenario, where module interactions are the most complex and operational constraints are the strongest. As reported in \hyperref[tab:performance]{Table~\ref*{tab:performance}}, although Qwen3-Coder-480B-A35B-Instruct and DeepSeek-V3 (685B) have substantially larger parameter scales, their success rates are only 51.40\% and 50.93\%, respectively, both markedly lower than the 83.49\% achieved by GLM-4.7 (358B). This result indicates that performance in complex scheduling tasks is not determined solely by parameter scale. Although larger models possess richer prior knowledge, they are also more likely to exhibit overly broad knowledge invocation and divergent reasoning paths under complex emergency situations, which makes it difficult to rapidly localize the key functional modules genuinely affected by the emergency situations. When localization fails, the decisions generated by global reasoning are more likely to deviate from the internal constraints and execution paths of the system, which ultimately reduces the overall task success rate. In contrast, MAFIG no longer performs reasoning and global modification over the entire scheduling system. Instead, it confines repair to the affected atomic functions and achieves rapid functional adjustment through targeted repair. Benefiting from this mechanism, MAFIG attains a success rate of 97.50\% in the deck scenario, with an average processing time of only 0.19 s. These results demonstrate that, compared with approaches relying solely on C-LLMs with massive parameter scales, the localized repair paradigm of MAFIG can more effectively suppress the risk of Execution Misalignment in complex decoupled scheduling systems, while preserving both decision-making accuracy and response efficiency.

\subsection{Effectiveness Evaluation of the MAFIG Architecture }
To evaluate the effectiveness of the MAFIG architecture, we select three open-source small models as the backbones of the Perception Agent and the Emergency Decision Agent, namely Qwen2.5-Coder-7B, Meta-Llama-3.1-8B-Instruct, and GLM-4-9B-Chat. In this experiment, the original small models are directly integrated into MAFIG, so that the contribution of the architecture itself can be examined independently of the proposed distillation mechanism. As reported in \hyperref[tab:ablation_sfl]{Table~\ref*{tab:ablation_sfl}}, three lightweight models achieve substantial improvements in task success rate after deployment in MAFIG across the port, warehouse, and deck scenarios, while the average processing time is also significantly reduced. In contrast, the original small models exhibit generally low task success rates in the three scenarios, and their performance declines further as scenario complexity increases. For example, in the port scenario, the success rates of Qwen2.5-Coder-7B and GLM-4-9B-Chat are only 25.62\% and 23.62\%, respectively. In the more complex deck scenario, the success rates of Qwen2.5-Coder-7B and Meta-Llama-3.1-8B-Instruct further decrease to 17.45\% and 10.90\%. These results indicate that, when confronted with emergency situations, the original models must infer and revise the complete scheduling logic directly from the emergency description. This setting is easily affected by the broad task scope, the numerous constraints, and the complex generation space, which makes satisfactory decision-making difficult.

In contrast, MAFIG transforms the originally complex problem of global scheduling reconstruction into a repair problem over local functional modules. It no longer requires the model to perform reasoning and global modifications over the entire scheduling system, which effectively reduces task difficulty and improves the task success rate of the original small models. In the warehouse scenario, the success rate of GLM-4-9B-Chat increases from 36.18\% to 69.10\%. Notably, in the most complex deck scenario, the success rate of Qwen2.5-Coder-7B still improves from 17.45\% to 48.91\%.

Beyond the improvement in success rate, MAFIG also enhances task processing efficiency. As shown in \hyperref[tab:ablation_sfl]{Table~\ref*{tab:ablation_sfl}}, in the deck scenario, the average processing time of Qwen2.5-Coder-7B is reduced from 0.39\,s to 0.16\,s, while Meta-Llama-3.1-8B-Instruct decreases from 0.70\,s to 0.12\,s. This improvement arises because MAFIG confines the decision scope to the affected local atomic functions, which substantially reduces the additional reasoning overhead introduced by irrelevant context and therefore shortens the overall processing time for emergency handling.

\begin{table*}[htbp]
    \centering
    \caption{Evaluating the Effect of MAFIG Architecture on Backbone Model Performance.}
    \label{tab:ablation_sfl}
    \begin{tabular*}{\textwidth}{@{\extracolsep{\fill}} c l c c c c @{}}
        \toprule
        \textbf{Task} & \textbf{Model} & \textbf{Method} & \textbf{Total Time (s)} & \textbf{Avg Time (s)} & \textbf{Success Rate} \\
        \midrule
        
        % Port 部分
        \multirow{6}{*}{Port} 
        & Qwen2.5-Coder-7B & --- & 69.20 & 0.35 & 25.62\% \\
        & Meta-Llama-3.1-8B-Instruct & --- & 101.39 & 0.51 & 30.15\% \\
        & GLM-4-9B-Chat & --- & 94.89 & 0.48 & 23.62\% \\
        & Qwen2.5-Coder-7B & MAFIG(w/o SFL) & 63.19 & \textbf{0.32} & 62.81\% \\
        & Meta-Llama-3.1-8B-Instruct & MAFIG(w/o SFL) & 71.06 & 0.36 & 48.24\% \\
        & GLM-4-9B-Chat & MAFIG(w/o SFL) & 86.11 & 0.43 & \textbf{64.82\%} \\
        \cmidrule{2-6}
        
        % Warehousing 部分
        \multirow{6}{*}{Warehousing} 
        & Qwen2.5-Coder-7B & --- & 151.18 & 0.38 & 25.87\% \\
        & Meta-Llama-3.1-8B-Instruct & --- & 407.56 & 1.02 & 30.15\% \\
        & GLM-4-9B-Chat & --- & 186.05 & 0.47 & 36.18\% \\
        & Qwen2.5-Coder-7B & MAFIG(w/o SFL) & 87.87 & \textbf{0.22} & 55.53\% \\
        & Meta-Llama-3.1-8B-Instruct & MAFIG(w/o SFL) & 101.55 & 0.26 & 40.45\% \\
        & GLM-4-9B-Chat & MAFIG(w/o SFL) & 136.39 & 0.34 & \textbf{69.10\%} \\
        \cmidrule{2-6}
        
        % Deck 部分
        \multirow{6}{*}{Deck} 
        & Qwen2.5-Coder-7B & --- & 250.13 & 0.39 & 17.45\% \\
        & Meta-Llama-3.1-8B-Instruct & --- & 449.40 & 0.70 & 10.90\% \\
        & GLM-4-9B-Chat & --- & 328.78 & 0.51 & 24.45\% \\
        & Qwen2.5-Coder-7B & MAFIG(w/o SFL) & 103.97 & 0.16 & \textbf{48.91\%} \\
        & Meta-Llama-3.1-8B-Instruct & MAFIG(w/o SFL) & 79.66 & \textbf{0.12} & 31.35\% \\
        & GLM-4-9B-Chat & MAFIG(w/o SFL) & 125.01 & 0.19 & 42.37\% \\
        
        \bottomrule
    \end{tabular*}
\end{table*}
\begin{table*}[htbp]
    \centering
    \caption{Latency and Success Rate Analysis of MAFIG With Versus Without SFL Across Different Tasks.}
    \label{tab:time_accuracy}
    \resizebox{\textwidth}{!}{%
    \begin{tabular}{c c c c c c c c}
        \toprule
        \textbf{Task} & \textbf{Model} & \textbf{Method} & \textbf{Perception Time (s)} & \textbf{Decision Time (s)} & \textbf{Total Time (s)} & \textbf{Avg Time (s)} & \textbf{Accuracy} \\
        \midrule
        
        % Port 部分
        \multirow{3}{*}{Port} 
        & \multirow{3}{*}{Qwen2.5-Coder-7B}
        & --- & --- & --- & 69.20 & 0.35 & 25.62\% \\
        & & MAFIG(w/o SFL) & 18.25 & 44.94 & 63.19 & 0.32 & 62.81\% \\
        & & \textbf{MAFIG} & \textbf{18.63} & \textbf{46.35} & \textbf{64.98} & \textbf{0.33} & \textbf{98.49\%} \\
        \midrule
        
        % Warehousing 部分
        \multirow{3}{*}{Warehousing} 
        & \multirow{3}{*}{Qwen2.5-Coder-7B}
        & --- & --- & --- & 151.18 & 0.38 & 25.87\% \\
        & & MAFIG(w/o SFL) & 23.02 & 64.85 & 87.87 & 0.22 & 55.53\% \\
        & & \textbf{MAFIG} & \textbf{22.08} & \textbf{68.19} & \textbf{90.27} & \textbf{0.23} & \textbf{94.97\%} \\
        \midrule
        
        % Deck 部分
        \multirow{3}{*}{Deck} 
        & \multirow{3}{*}{Qwen2.5-Coder-7B}
        & --- & --- & --- & 250.13 & 0.39 & 17.45\% \\
        & & MAFIG(w/o SFL) & 31.30 & 72.67 & 103.97 & 0.16 & 48.91\% \\
        & & \textbf{MAFIG} & \textbf{31.40} & \textbf{88.47} & \textbf{119.87} & \textbf{0.19} & \textbf{97.50\%} \\
        
        \bottomrule
    \end{tabular}%
    }
\end{table*}

\subsection{Effectiveness Evaluation of SFL}

Building on the preceding validation of the MAFIG architecture, this section further examines the contribution of SFL to the overall performance of the proposed method. Specifically, we compare MAFIG with MAFIG (w/o SFL) in terms of task success rate and processing efficiency across the three scenarios, and report the corresponding results in \hyperref[tab:time_accuracy]{Table~\ref*{tab:time_accuracy}}. Across the three scenarios, MAFIG (w/o SFL) achieves success rates of 62.81\%, 55.53\%, and 48.91\%, respectively. Although the MAFIG architecture yields substantial performance gains for small models, the knowledge capacity, reasoning depth, and capabilities of these small models in complex decision tasks remain limited by their parameter scale. After applying SFL, the success rates rise further to 98.49\%, 94.97\%, and 97.50\%, respectively. These results indicate that SFL can effectively transfer the knowledge and decision patterns exhibited by C-LLMs to small models during emergency understanding, logical reasoning, and decision generation. As a result, the lightweight local models acquire stronger decision-making capability and achieve higher repair accuracy under complex constraints.

Notably, although the deck scenario involves higher task complexity and has only 120 training samples, fewer than the 170 samples used in the warehouse scenario, MAFIG still achieves a success rate of 97.50\%, exceeding the 94.97\% achieved in the warehouse scenario. This result indicates that distillation effectiveness is not determined solely by the number of training samples, but is also closely related to the distribution characteristics of emergency situations across scenarios. As shown in \hyperref[FIG:5]{Fig.~\ref*{FIG:5}}, the deck scenario contains a greater variety of emergency situations than the warehouse scenario. Nevertheless, the corresponding decision adjustments are typically concentrated on several critical functional modules. This characteristic facilitates the learning of representative handling patterns. Furthermore, SFL enables the small model to more effectively learn the decision patterns exhibited by large models in these recurrent critical cases, which explains why high accuracy is preserved even in the more complex scenario.

\begin{figure*}
	\centering
	\includegraphics[width=.9\textwidth]{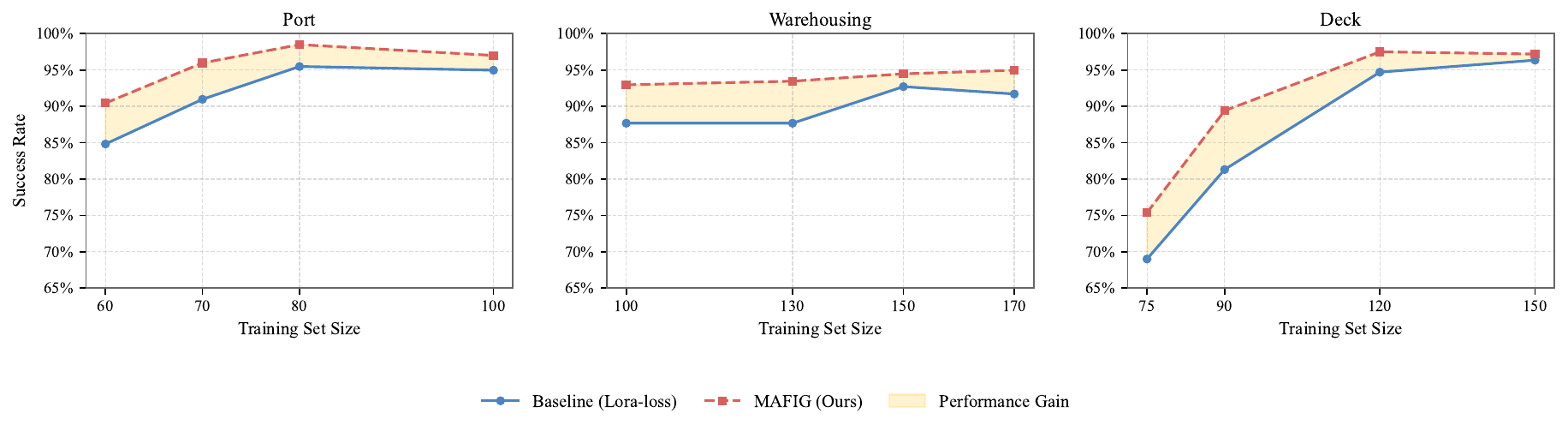}
	\caption{Performance comparison of SFL and LoRA under different training set sizes across tasks.}
	\label{FIG:6}
\end{figure*}

\begin{figure*}[t]
	\centering
	\includegraphics[width=.9\textwidth]{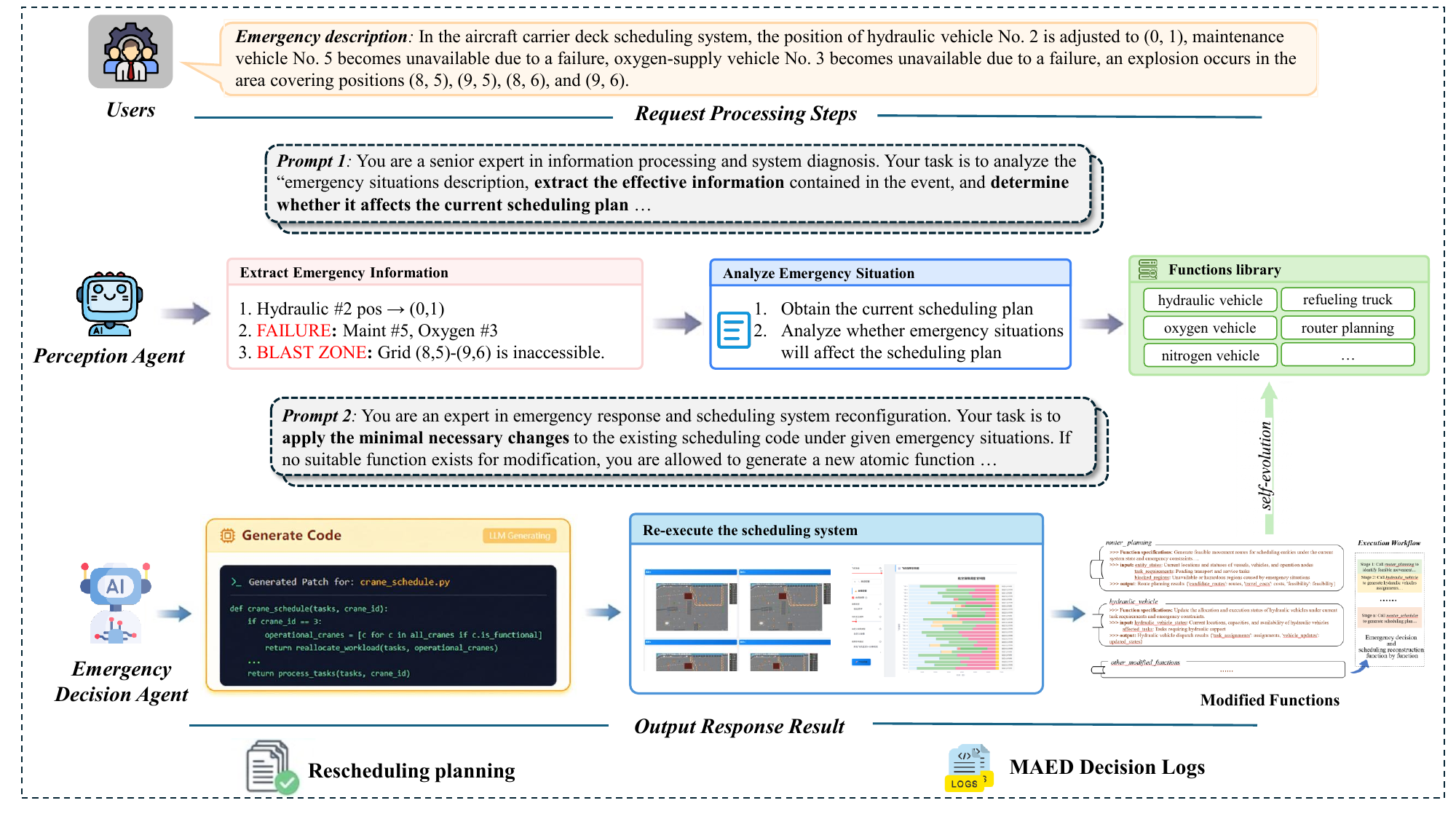}
	\caption{Case study of MAFIG in the deck scheduling scenario under concurrent emergency situations.}
	\label{FIG:7}
\end{figure*}

\subsection{Scenario Applicability Analysis of SFL}

To further evaluate the effectiveness of the proposed SFL in emergency decision tasks, we construct training sets of different sizes with multiple categories of emergency situations across the port, warehouse and deck scenarios. We select Qwen2.5-Coder-7B as the backbone model to compare SFL with conventional LoRA fine-tuning.

As shown in \hyperref[FIG:6]{Fig.~\ref*{FIG:6}}, under limited training data, conventional LoRA achieves only modest performance gains, whereas SFL exhibits more stable and consistent improvements. In the port scenario, when the training set contains only 60 samples, LoRA achieves a success rate of 84.82\%, while SFL already reaches 90.45\%. When the training set is increased to 70 samples, the success rate of SFL further rises to 95.97\%, which remains clearly higher than the 90.97\% obtained by LoRA. This result indicates that, under limited training data, SFL explicitly identifies the true edited regions through a diff-based strategy and reinforces the supervision signal on critical modification segments with a span-weighted cross-entropy loss function, enabling the model to concentrate on the code adjustment patterns that are genuinely required for emergency handling. In contrast, conventional LoRA typically optimizes the entire output sequence in a uniform manner. In emergency scheduling reconstruction tasks, the actual learning target is often confined to a local modification region within a function, while unchanged contextual code still accounts for the majority of the sequence. Consequently, LoRA tends to disperse parameter updates over irrelevant background code, which weakens learning of the core modification logic.

As scenario complexity increases, the advantage of SFL becomes more pronounced. In the warehouse scenario, when the training set contains 100 samples, SFL improves the success rate from 87.68\% under LoRA to 92.96\%. This result shows that SFL can still maintain stable performance gains in scenarios with more complex interactions and more numerous constraints. The underlying reason is that, in complex scenarios with diverse emergency situations, SFL is more suited to concentrate learning on the effective modification regions. If the full-sequence average optimization used by LoRA is retained, the model is more easily affected by redundant contextual information. A similar trend can also be observed in the deck scenario. When the training set contains only 75 samples, SFL achieves a success rate of 75.38\%, exceeding the 69.00\% obtained by LoRA. When the training set is increased to 90 samples, the success rate of SFL further rises to 89.40\%, whereas LoRA reaches only 81.31\%.

\subsection{Case Study}
To further validate the practical effectiveness of MAFIG under highly dynamic scenarios with multiple constraints, we conduct a case study on deck scheduling, which is the most structurally complex of the three scenarios. The case study shows that MAFIG can precisely localize the affected atomic functions and perform code repair under multiple concurrent emergency situations, which reduces the risk of Execution Misalignment in complex decoupled systems.

As illustrated in \hyperref[FIG:7]{Fig.~\ref*{FIG:7}}, the system receives the following unstructured emergency description: ``Hydraulic vehicle No.~2 is adjusted to $(0,1)$, maintenance vehicle No.~5 and oxygen supply vehicle No.~3 become unavailable due to failures, and an explosion occurs in the grid region spanning $(8,5)$ to $(9,6)$.'' Upon receiving these instructions, MAFIG invokes the Perception Agent to perform semantic parsing, extract heterogeneous emergency factors, and assess their substantive impact on the current scheduling plan. If the current plan is determined to be affected, the Perception Agent rapidly localizes the relevant functional modules in the function library and transfers them, along with the emergency information, to the Emergency Decision Agent. The Emergency Decision Agent then applies the minimum necessary modifications to the relevant core code. Finally, the system produces an executable rescheduling solution and adds the validated atomic functions to the atomic function library, which repairs the scheduling logic and supports continued adaptive evolution.

\section{Conclusion}
In this paper, we propose the Multi-agent Driven Formal Instruction Generation Framework (MAFIG), which constrains repair to locally affected atomic functions. The framework comprises a Perception Agent and an Emergency Decision Agent, and SFL transfers the emergency handling capability of Cloud Large Language Models to lightweight local models, reducing response time in emergency situations while preserving high decision-making quality. Simulation experiments in port, warehouse, and deck scheduling scenarios demonstrate that MAFIG consistently outperforms multiple powerful Cloud Large Language Models.

\section*{Acknowledgments}
This work is supported by the National Natural Science Foundation of China (Grant Nos.62325602, 62506342), Henan Provincial Natural Science Foundation Youth Category B Project (Grant No.262300421217), Postgraduate Education Research Project of Zhengzhou University (Grant No.YJSJY2025138), the China Postdoctoral Science Foundation (Grant No.2025M781527).

\end{document}